\documentclass[conference]{IEEEtran}
\IEEEoverridecommandlockouts
\usepackage{cite}
\usepackage{amsmath,amssymb,amsfonts}
\usepackage{algorithmic}
\usepackage{graphicx}
\usepackage{textcomp}
\usepackage{tabularray}
\usepackage[table,xcdraw]{xcolor}
\usepackage{enumitem}
\usepackage{hyperref}
\usepackage{lipsum}
\usepackage{balance}
\usepackage{url}

\def\BibTeX{{\rm B\kern-.05em{\sc i\kern-.025em b}\kern-.08em
    T\kern-.1667em\lower.7ex\hbox{E}\kern-.125emX}}
\begin{document}

\title{State of play and future directions in industrial computer vision AI standards
\thanks{The research leading to these results has received funding from the European Union’s Horizon Europe research and innovation programme under grant agreement No 101073876 (Ceasefire). This publication reflects only the authors views. The European Union is not liable for any use that may be made of the information contained therein.}

\author{\IEEEauthorblockN{Artemis Stefanidou}
\IEEEauthorblockA{\textit{Dept. of Informatics and Telematics} \\
\textit{Harokopio University of Athens}\\
\textit{Athens, Greece}\\
astefanidou@hua.gr}
\and
\IEEEauthorblockN{Panagiotis Radoglou-Grammatikis}
\IEEEauthorblockA{\textit{K3Y Ltd.}\\ 
\textit{Sofia, Bulgaria}\\
pradoglou@k3y.bg}
\and
\IEEEauthorblockN{Vasileios Argyriou}
\IEEEauthorblockA{\textit{Dept. of Networks and Digital Media}\\
\textit{Kingston University}\\
\textit{Kingston upon Thames, United Kingdom}\\
vasileios.argyriou@kingston.ac.uk}
\and
\IEEEauthorblockN{Panagiotis Sarigiannidis}
\IEEEauthorblockA{\textit{Dept. of Electrical and Computer Engineering}\\ 
\textit{University of Western Macedonia}\\
\textit{Kozani, Greece}\\
psarigiannidis@uowm.gr}
\and
\IEEEauthorblockN{Iraklis Varlamis}
\IEEEauthorblockA{\textit{Dept. of Informatics and Telematics} \\
\textit{Harokopio University of Athens}\\
\textit{Athens, Greece}\\
varlamis@hua.gr}
\and
\IEEEauthorblockN{Georgios Th. Papadopoulos}
\IEEEauthorblockA{\textit{Dept. of Informatics and Telematics} \\
\textit{Harokopio University of Athens}\\
\textit{Athens, Greece}\\
g.th.papadopoulos@hua.gr}
}
}

\maketitle

\begin{abstract}
The recent tremendous advancements in the areas of Artificial Intelligence (AI) and Deep Learning (DL) have also resulted into corresponding remarkable progress in the field of Computer Vision (CV), showcasing robust technological solutions in a wide range of application sectors of high industrial interest (e.g., healthcare, autonomous driving, automation, etc.). Despite the outstanding performance of CV systems in specific domains, their development and exploitation at industrial-scale necessitates, among other, the addressing of requirements related to the reliability, transparency, trustworthiness, security, safety, and robustness of the developed AI models. The latter raises the imperative need for the development of efficient, comprehensive and widely-adopted industrial standards. In this context, this study investigates the current state of play regarding the development of industrial computer vision AI standards, emphasizing on critical aspects, like model interpretability, data quality, and regulatory compliance. In particular, a systematic analysis of launched and currently developing CV standards, proposed by the main international standardization bodies (e.g. ISO/IEC, IEEE, DIN, etc.) is performed. The latter is complemented by a comprehensive discussion on the current challenges and future directions observed in this regularization endeavor.
\end{abstract}


\section{Introduction}
The recent outstanding advancements in the areas of Artificial Intelligence (AI) \cite{wang2023scientific} and Machine Learning (ML) \cite{sarker2021machine}, largely due to the merits introduced by the so called Deep Learning (DL) \cite{dong2021survey} paradigm, have also boosted tremendous achievements in the field of Computer Vision (CV). In particular, DL methodologies rely on the use of large-scale Deep Neural Network (DNN) architectures \cite{chai2021deep}. Such DNN-based have exhibited unprecedented performance in a wide set of visual understanding tasks, clearly surpassing similar hand-crafted approaches by a large margin. It needs to be highlighted though that the fundamental prerequisite for this eminent visual interpretation and reasoning capability is the availability of ever larger and sufficiently diverse training datasets \cite{shrestha2019review, alimisis2025advances}.

CV methods have received particular attention and have introduced increased potentials in a wide set of application sectors of far-reaching societal and economic impact, demonstrating efficient, robust and reliable technological solutions. Specifically, CV techniques have been successfully deployed in the following, among numerous other, industries: a) Manufacturing \cite{zhou2022computer}, b) Healthcare \cite{esteva2021deep, konstantakos2025self}, c) Agriculture \cite{dhanya2022deep}, d) Automation \cite{papadopoulos2021towards}, e) Transportation \cite{liu2021smart}, f) Security \cite{mademlis2024invisible}, g) Education \cite{joshi2023towards}, h) Retail \cite{shoman2022region}, i) Robotics \cite{papadopoulos2022user}, j) Sports \cite{cui2023sportsmot}, etc.

Despite the outstanding achievements and contributions of CV systems in multiple sectors, their deployment at industrial-scale has raised, apart from handling strictly algorithmic and technical obstacles, several challenges regarding aspects related to, among others, the reliability, transparency, explainability, interpretability, fairness, trustworthiness, security, safety, and robustness of the developed AI models \cite{li2023trustworthy, rodis2024multimodal}. This fact inevitably raises the urgent demand for the development of efficient, comprehensive and widely-adopted relevant industrial standards. The latter need to handle and regulate all stages of the CV module development life-cycle, ranging from the collection of training data and the development of the actual software modules to aspects related to the usage, deployment, evaluation and adoption of the generated solutions.  

In this paper, the current state of play regarding the development of industrial computer vision AI standards is systematically investigated, putting particular focus on examining crucial aspects, like model interpretability, data quality, and regulatory compliance. The main contributions of this study are as follows:
\begin{itemize} 
    \item The specific steps involved in the AI standard development process and previous standards' examination works are discussed in detail.
    \item A thorough and systematic analysis of launched and currently developing CV standards, proposed by the main international standardization bodies (e.g. ISO/IEC, IEEE, DIN, etc.) is performed.    
    \item A comprehensive discussion on the current challenges and future directions observed in the industrial CV standard regularization endeavor is provided.
\end{itemize}

The remainder of the article is organized as follows: Section \ref{sec:standardDevelopmentProcess} discusses the AI standard development process and previous standards' examination studies. Section \ref{sec:standards} systematically analyzes the currently available industrial computer vision AI standards. Section \ref{sec:challenges} details current challenges and future development directions. Section \ref{sec:conclusion} concludes the paper.

\section{AI standard development process and previous work}
\label{sec:standardDevelopmentProcess}

Standards are crucial for ensuring consistency, interoperability, and quality control across multiple technological domains, including AI. A standard comprises a formally established guideline that outlines specifications, procedures, and/or criteria to achieve specific results. Such formalized guidelines promote reliability, safety, and compliance, while also fostering innovation by providing a structured framework for development and evaluation.

The creation of a standard typically involves multiple stakeholders, including industry leaders, regulatory bodies, researchers, and governmental organizations. This process generally consists of the following main steps\cite{ISO/Directives2024, IEEEstandards2022}:

\begin{itemize} 
    \item \underline{Proposal}: Identification of the need for a new standard, often driven by emerging challenges or technological gaps. 
    \item \underline{Development}: Experts collaborate to draft specifications, considering best practices and technological feasibility. 
    \item \underline{Review and consensus}: Stakeholder feedback is collected and refinements are made to reach a consensus. 
    \item \underline{Approval and publication}: A governing body, such as ISO or IEEE, formally adopts/approves the proposed standard. 
    \item \underline{Implementation and maintenance}: The standard is adopted/used by organizations and is periodically updated so as to remain relevant to technological advancements. 
\end{itemize}

Given the fact that AI-based systems continuously become more sophisticated and embedded across various sectors, ensuring their quality and compliance with regulatory frameworks remains a key priority. In this context, best practices from the software engineering field can be initially adapted to the case of AI system development, in order to create systems that are high-performing, transparent, explainable, and resilient. However, in order to address the ever growing complexity and sophisticated nature of AI systems, a combination of regulatory frameworks, engineering principles, and rigorous evaluation metrics is required. The latter dictates the need for well-defined standards to ensure AI systems are not only innovative, but also trustworthy and regulation compliant.

Concerning AI-driven computer vision systems, these rely heavily on existing research and regulatory frameworks, in order to foster quality assurance. Continuously taking into account advancements in AI methodologies, along with compliance to industrial standards, is crucial component for ensuring accuracy and reliability of such systems. Oviedo et al. \cite{oviedo2024iso} perform a broad review of AI standards (mainly those defined by ISO/IEC) with a particular focus on software aspects, namely at the level of process and product quality, and at the level of data quality of applications integrating AI systems. Additionally, Vyas and Xu \cite{vyas2024key} investigate AI solutions in Advanced Driver Assistance Systems (ADAS), showcasing how perception, sensor fusion, and deep learning contribute to precision in defect detection and automation. Moreover, Gültekin-Várkonyi \cite{gultekin2024navigating} highlights the challenges posed by facial recognition technology, especially in relation to privacy risks and the importance of regulatory compliance. This underscores the necessity for clear technical specifications, data quality, and transparent terminology, enabling public understanding and adoption.

\begin{table*}[t]
    \centering
    \caption{Industrial computer vision AI standards\\(Topic: `Accuracy, Performance, and Processing' (APP), `System Design and Architecture' (SDA), `Security, Robustness, and Risk management' (SRR), `Data Management and Quality' (DMQ), `Ethics, Privacy, and Fairness' (EPF), and `Biometric Identification and Authentication' (BIA))}
    \begin{tblr}{
      colspec={X[1.8,l] X[1.2,c] X[1.6,c] X[0.8,c] X[0.3,c] X[0.3,c] X[0.3,c] X[0.3,c] X[0.3,c] X[0.3,c]}, 
      row{1} = {font=\bfseries, c, bg=gray!70},
      row{3,5,7,9,11,13,15,17,19,21,23} = {bg=gray!30},
      row{even} = {bg=gray!20}, 
      hlines, 
      vlines 
    }
        \SetCell[r=2]{c} \textbf{Standard} & 
        \SetCell[r=2]{c} \textbf{Organization} & 
        \SetCell[r=2]{c} \textbf{Domain} & 
        \SetCell[r=2]{c} \textbf{Year} & 
        \SetCell[c=6]{c} \textbf{Topic} & \\ 
        
        & & & &APP&SDA&SRR&DMQ&EPF&BIA\\
        
        ISO/IEC DIS 9868 \cite{ISO/IECDIS9868} & ISO/IEC & Horizontal & 2025 & \checkmark & & \checkmark & & & \\
        ISO/IEC WD TS 24358 \cite{ISO/IECWDTS24358} & ISO/IEC & Horizontal & 2025 &  \checkmark & & \checkmark & & \checkmark \\
        ISO/IEC DIS 19795-10 \cite{ISO/IECDIS19795-10} & ISO/IEC & Horizontal & 2024 &  \checkmark & & & & \checkmark \\
        ISO/IEC CD 29794-5.3 \cite{ISO/IECCD29794-5.3} & ISO/IEC & Horizontal & 2024 &  & & & \checkmark & & \\
        ISO/IEC DTS 22604 \cite{ISO/IECDTS22604} & ISO/IEC & Horizontal & 2024 &  & & & & & \checkmark \\
        ISO/IEC TR 24741 \cite{ISO/IECTR24741:2018} & ISO/IEC & Horizontal & 2024 &  \checkmark & \checkmark & \checkmark & & \checkmark \\
        ISO/IEC DIS 5152 \cite{ISO/IECDIS5152} & ISO/IEC & Horizontal & 2024 &  \checkmark & & & & \\
        ISO/IEC TR 20322 \cite{ISO/IECTR20322:2023} & ISO/IEC & Horizontal & 2023 &  & & & & \checkmark \\
        ISO/IEC TR 24714-1 \cite{ISO/IECTR24714-1} & ISO/IEC & Horizontal & 2023 &  \checkmark & \checkmark & \checkmark & & \checkmark  \\
        ISO/IEC TR 22116 \cite{ISO/IECTR22116:2021} & ISO/IEC & Horizontal & 2021 &  \checkmark & & & & \checkmark  \\
        ISO/TR 24291 \cite{ISO/TR24291:2021} & ISO & Healthcare and medicine & 2021 &  \checkmark & & & \checkmark &  \\
        IEEE 3110-2025 \cite{IEEE3110-2025} & IEEE & Horizontal & 2025 &  & \checkmark & & &  \\
        IEEE 3161 \cite{IEEE3161} & IEEE & Horizontal & 2023 &  \checkmark & \checkmark & & &  \\
        IEEE 3129-2023 \cite{IEEE3129-2023} & IEEE & Horizontal & 2023 &  & & \checkmark & & \\
        IEEE 2945-2023 \cite{IEEE2945-2023} & IEEE & Horizontal & 2023 &  \checkmark & \checkmark & & & \checkmark \\
        P3157 \cite{P3157} & IEEE & Horizontal & 2022 &  & & \checkmark & &  \\
        IEEE 2671-2022 \cite{IEEE2671-2022} & IEEE & Manufacturing & 2022 &  \checkmark & \checkmark & & &  \\
        DIN SPEC 13266 \cite{DINSPEC13266} & DIN & Horizontal & 2020 &  \checkmark & & \checkmark & \checkmark &  \\
        DIN SPEC 13288 \cite{DINSPEC13288} & DIN & Healthcare and medicine & 2021 &  \checkmark & & \checkmark & & \checkmark  \\
        JT021025 \cite{JT021025} & CEN, CENELEC & Horizontal & 2025 &  \checkmark & & & &  \\
        BS 9347:2024 \cite{BS9347:2024} & BSI & Defense and security & 2024 &  & & & & \checkmark  \\
    \end{tblr}
    \label{tab:ai_standards_computer_vision}
\end{table*}

\section{Industrial computer vision AI standards}
\label{sec:standards}

The development of industrial AI standards for computer vision has received increased attention world-wide, with several key organizations contributing to the establishment of technical frameworks and guidelines. In Particular, the `Canadian Standards Association Group' (CSA Group)\footnote{\href{https://www.csagroup.org/about-csa-group/}{Canadian Standards Association Group}} plays a pivotal role in developing standards for emerging technologies, including AI. In parallel, the `Innovation, Science and Economic Development Canada' (ISED)\footnote{\href{http://ised-isde.canada.ca/site/ised/en}{Innovation, Science and Economic Development Canada}} institution coordinates efforts to establish regulatory frameworks for AI governance. Additionally, the `Standards Council of Canada' has published the `Canadian Data Governance Standardization Roadmap'\footnote{\href{https://scc-ccn.ca/resources/publications/canadian-data-governance-standardization-roadmap}{Canadian Data Governance Standardization Roadmap}} that focuses on issues such as data governance, privacy protection, transparency, and reliability of AI systems. 
On the other hand, the US `National Institute of Standards and Technology' (NIST)\footnote{\href{https://www.nist.gov/}{National Institute of Standards and Technology}} has developed the `Artificial Intelligence Risk Management Framework 1.0' (AI RMF 1.0)\footnote{\href{https://nvlpubs.nist.gov/nistpubs/ai/nist.ai.100-1.pdf}{Artificial Intelligence Risk Management Framework 1.0}} for providing essential guidelines for managing risks associated with AI technologies, as well as `A Plan for Global Engagement on AI Standards'\footnote{\href{https://nvlpubs.nist.gov/nistpubs/ai/NIST.AI.100-5.pdf}{A Plan for Global Engagement on AI Standards}} to drive the development of AI-related consensus standards. Moreover, the European Commission coordinates the establishment of legal and ethical frameworks for AI, through initiatives like the `AI Act'\footnote{\href{https://digital-strategy.ec.europa.eu/en/policies/regulatory-framework-ai}{AI Act}} regulation framework. Similarly, technical standards are developed by organizations such as the `European Committee for Standardization' (CEN) and the `European Telecommunications Standards Institute' (ETSI). All the above mentioned coordinated efforts involve activities that aim to ensure the consistent, secure, and ethical development and deployment of AI computer vision technologies.

The main organizations driving the development of industrial
computer vision AI standards are the `International Organization for Standardization' (ISO), the `International Electrotechnical Commission' (IEC), the `Institute of Electrical and Electronics Engineers Standards Association' (IEEE), the `German Institute for Standardization' (DIN), the `European Committee for Standardization' (CEN), the `European Committee for Electrotechnical Standardization' (CENELEC), and the `British Standards Institution' (BSI).

\begin{figure*}[t]
    \centering
    \includegraphics[width=0.9\linewidth]{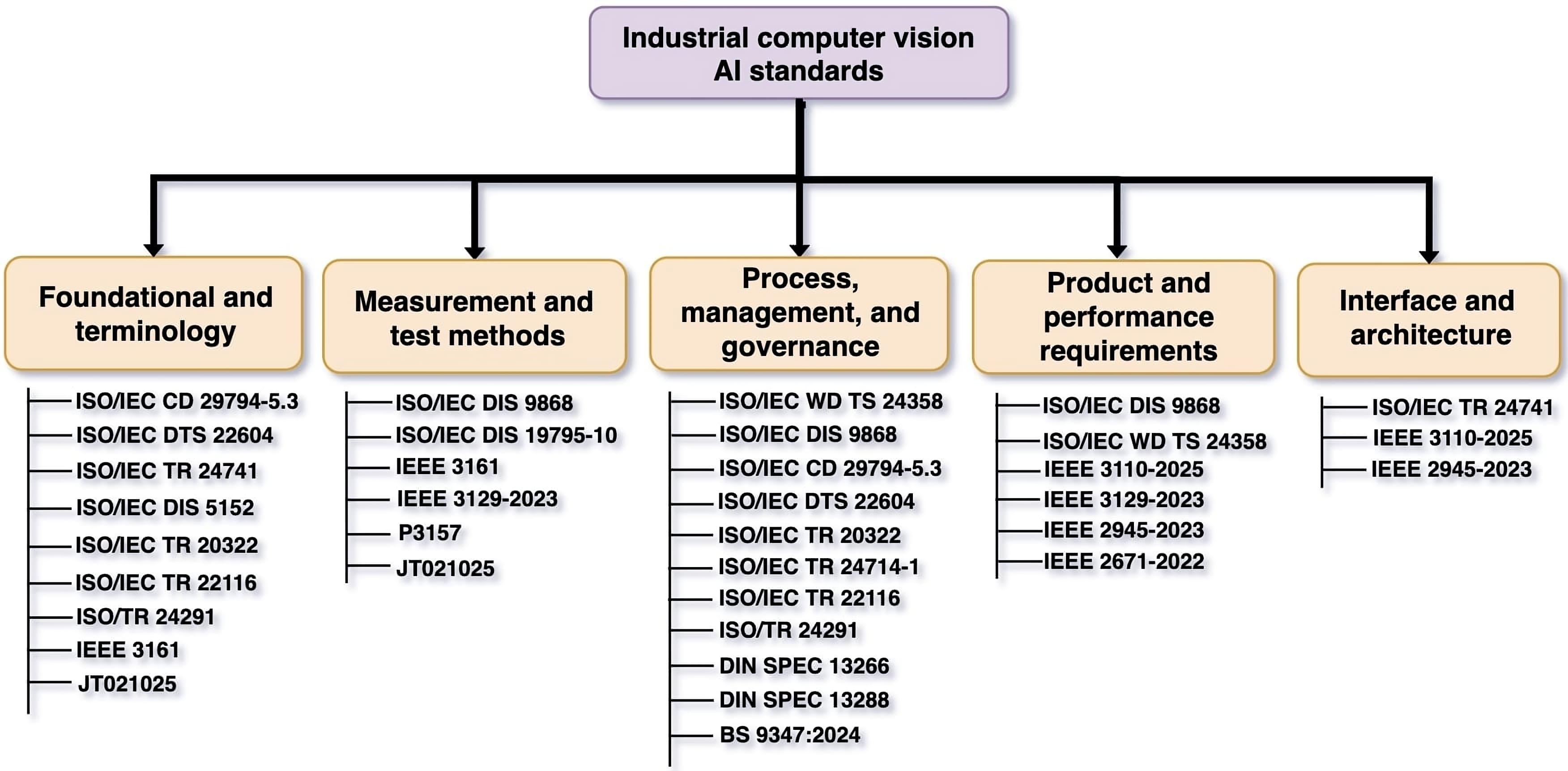}
    \caption{Industrial computer vision AI standards}
    \label{f:diagramCateg}
\end{figure*}

Table \ref{tab:ai_standards_computer_vision} illustrates the main industrial CV AI standards currently available, where for each record the establishing organization, the application domain, the year of publication and the particular topic that the respective standard refers to are provided. Regarding the standards' application domain, the following main ones have been considered so far:
\begin{itemize} 
    \item \underline{Horizontal}: This includes standards that investigate fundamental AI system properties that are tailored to multiple and diverse sectors, aiming at ensuring broad applicability, consistency, and interoperability. 
    \item \underline{Healthcare and medicine}: This comprises standards that ensure that AI technologies, such as medical imaging and diagnostic systems, meet high accuracy and safety requirements, in order to support patient health management, while complying with relevant healthcare regulations.
    \item \underline{Manufacturing}: Standards under this category focus on enhancing automation, quality control, and defect detection using AI-powered vision systems, targeting to enhance reliability and efficiency.
    \item \underline{Defense and security}: This incorporates standards that address privacy/ethical concerns and security risks, ensuring responsible and secure deployment of AI technologies, like facial recognition and surveillance systems. 
\end{itemize}
Additionally, the available standards cover a broad range of topics, including the following main ones:
\begin{itemize}
    \item \underline{Accuracy, Performance, and Processing (APP)}: This ensures that AI models produce reliable and consistent results, optimizing efficiency and computational requirements. It also covers performance metrics, such as precision, recall, and latency.
    \item \underline{System Design and Architecture (SDA)}: This focuses on the structural organization of AI systems, including model deployment, hardware-software integration, and scalability aspects.
    \item \underline{Security, Robustness, and Risk management (SRR)}: This covers safeguards against adversarial attacks, system vulnerabilities, and risk mitigation strategies, in order to enhance AI reliability in critical applications.
    \item \underline{Data Management and Quality (DMQ)}: This focuses on data integrity, consistency, and standardization, in order to ensure high-quality training datasets and to reduce bias in AI systems.
    \item \underline{Ethics, Privacy, and Fairness (EPF)}: This addresses biases in AI models, user privacy protection, and adherence to ethical AI principles, in order to ensure non-discriminatory and responsible AI deployment.
    \item \underline{Biometric Identification and Authentication (BIA)}: This establishes protocols for secure and ethical use of facial recognition and other biometric technologies, assuring compliance with legal and privacy regulations.
\end{itemize}

Taking into account the stage within the AI system development life-cycle, CV standards can be classified in the following key categories:
\begin{itemize}
    \item \underline{Foundational and terminology}: These standards define fundamental principles and basic terminology, ensuring consistency and clear communication across stakeholders.
    \item \underline{Measurement and test methods}: Such standards specify measurement methods and testing procedures, in order to assess reliability, accuracy, and quality aspects of AI systems.
    \item \underline{Process, management, and governance}: These provide guidelines for project management, regulatory compliance, and life-cycle oversight of AI systems.
    \item \underline{Product and performance requirements}: These define specifications for functionality, safety, and performance, ensuring systems achieve expected quality levels.
    \item \underline{Interface and architecture}: These outline system architecture design and integration, facilitating interoperability among components in computer vision AI systems.
\end{itemize}
The CV standards provided in Table \ref{tab:ai_standards_computer_vision} are classified to the above mentioned categories, as illustrated in Fig. \ref{f:diagramCateg}.

\section{Current challenges and future directions}
\label{sec:challenges}

Standardization efforts in the area of CV exhibit multiple challenges and increased potentials for future developments, stemming out from both the rapid technological advances in the field and factors like the fragmentation of standardization organizations, varying regulatory frameworks, and different ethical perspectives. Addressing these challenges will enable the establishment of more efficient and universally adopted guidelines, boosting in this way further proliferation and deployment of CV technologies. Key challenges and corresponding future directions are discussed in the followings.

\subsection{Fragmentation of standardization bodies and regulatory frameworks}
Currently, multiple (inter-)national bodies are involved in developing standards for CV, including organizations like ISO, IEEE, NIST, etc. Each such entity develops guidelines that often differ in scope and focus, leading to inconsistencies and contradictions in the produced guidelines. To make matters worse, the wide set of available (inter-)national regulatory frameworks and policies (e.g., EU AI Act, US AI governance strategies, etc.) introduce additional difficulties in achieving global convergence and consensus. In order to overcome these boundaries, closer and continuous collaboration among international organizations, governments, and private entities is needed.

\subsection{Bias and fairness in AI/CV systems}
Bias and fairness in AI remain significant concerns, as models trained on unbalanced datasets can lead to discriminatory outcomes. Although existing regulations stress the importance of bias mitigation, enforceable guidelines remain difficult to establish, due to differing ethical perspectives across different regions/states. More intense and deeper inter-national collaboration/initiatives can facilitate towards alleviating from this drawback.

\subsection{Security and privacy in AI/CV systems}
AI-based CV systems frequently process sensitive personal and industrial data, making them attractive targets for cyber-attacks, adversarial attacks, data breaches, and algorithmic manipulations. Despite the pressing need for standardized security protocols, no universally adopted solutions currently exist. Establishing globally recognized security frameworks, such as standardized encryption techniques, differential privacy methods and adversarial robustness measures, is essential to ensuring the safe deployment of AI-driven CV technologies.

\subsection{Thorough model evaluation}
Robustness and transparency in CV model evaluation is of paramount importance, as the diversity of assessment methodologies often leads to inconsistencies in performance reporting. Well-defined evaluation frameworks should be established to ensure traceability in training data, explainability of developed models and procedures, and enhancement of trust in AI-driven decision-making. Moreover, the establishment of standardized evaluation metrics and auditing mechanisms would enhance the credibility and reliability of the developed CV modules.

\subsection{Model risk assessment}
Given the rapid evolution of AI technologies, ongoing assessments are required to evaluate emerging risks. The current \textit{build-then-test} approach in AI development often neglects long-term risks associated with fine-tuning and post-deployment adaptation. In order to address this, comprehensive risk assessment frameworks need to be established, in order to continuously evaluate/monitor AI systems throughout their life-cycle. Incorporating human oversight, leveraging safety principles from specific industries (e.g., aviation and healthcare) and adopting real-time monitoring solutions would contribute to more robust AI governance.

\subsection{Robustness and transparency of AI/CV systems}
Enhancing robustness and explainability of AI models comprises a high priority, particularly in high-risk applications, such as facial recognition and autonomous navigation. Additionally, transparency remains a pressing issue, as many AI models operate as `black boxes', making it difficult to interpret their decision-making processes and to ensure accountability. Standardized testing methodologies should be implemented to assure that AI systems meet reliability and transparency requirements. Moreover, sector-specific standardization would be beneficial, since different sectors exhibit distinct operational constraints and ethical considerations. 

\subsection{Balancing with innovation potentials}
CV models need to be integrated with diverse hardware and software ecosystems, while excessive standardization risks to stifle innovation. Balancing regulatory oversight with the necessary flexibility for enabling technological advancements constitutes a critical point. In this respect, a well-regulated AI/CV ecosystem would not only promote responsible innovation, but would also safeguard against potential risks associated with uncontrollable AI developments.

\section{Conclusion}
\label{sec:conclusion}
Over the last years, Computer Vision (CV) has demonstrated tremendous advances and has introduced robust technological solutions in a wide range of application sectors of high industrial interest (e.g., healthcare, autonomous driving, automation, etc.). The deployment and exploitation of CV systems at industrial-scale urgently demands, among other, the addressing of requirements related to the reliability, transparency, trustworthiness, security, safety, and robustness of the developed AI models. The latter raises the imperative need for the establishment of efficient, comprehensive and widely-adopted industrial standards. This study investigated the current state of play regarding the development of industrial computer vision AI standards, emphasizing on critical aspects, like model interpretability, data quality, and regulatory compliance. Specifically, a thorough and systematic examination of launched and currently developing CV standards, proposed by the main international standardization bodies (e.g. ISO/IEC, IEEE, DIN, etc.) was provided. In a complementary way, a comprehensive discussion on the current challenges and future development directions in the field was performed.

\bibliographystyle{IEEEtran}
\balance
\bibliography{references}

\end{document}